\documentclass[10pt,twocolumn,letterpaper]{article}

\usepackage{cvpr}
\usepackage{times}
\usepackage{epsfig}
\usepackage{graphicx}
\usepackage{amsmath}
\usepackage{amssymb}
\usepackage{multirow}
\usepackage{booktabs}
\usepackage{array}
\usepackage{authblk}
\usepackage{footnote}
\usepackage{bm}
\makesavenoteenv{tabular}

\usepackage[breaklinks=true,bookmarks=false]{hyperref}

\cvprfinalcopy % *** Uncomment this line for the final submission

\def\cvprPaperID{****} % *** Enter the CVPR Paper ID here

\begin{document}

%%%%%%%%% TITLE
\title{The Solution for the AIGC Inference Performance Optimization Competition}
\author[1]{Sishun Pan}
\author[1]{Haonan Xu}
\author[1]{Zhonghua Wan}
\author[1]{{Yang Yang \thanks{Corresponding author: Yang Yang (yyang@njust.edu.cn)}}}
\affil[1]{Nanjing University of Science and Technology}

% \author{Haonan Xu$^{1}$, Yurui Huang$^{1}$, Sishun pan$^{1}$, Zhihao Guan$^{1}$, Yi Xu$^{2}$, Yang Yang$^{1,\ast}$ \\
% $^{1}$Nanjing University of Science and Technology\\
% $^{2}$Dalian University of Technology
\maketitle
%\thispagestyle{empty}
%%%%%%%%% ABSTRACT
\begin{abstract}
In recent years, the rapid advancement of large-scale pre-trained language models based on transformer architectures has revolutionized natural language processing tasks. Among these, ChatGPT has gained widespread popularity, demonstrating human-level conversational abilities and attracting over 100 million monthly users by late 2022. Concurrently, Baidu's commercial deployment of the Ernie Wenxin model has significantly enhanced marketing effectiveness through AI-driven technologies. This paper focuses on optimizing high-performance inference for Ernie models, emphasizing GPU acceleration and leveraging the Paddle inference framework. We employ techniques such as Faster Transformer for efficient model processing, embedding layer pruning to reduce computational overhead, and FP16 half-precision inference for enhanced computational efficiency. Additionally, our approach integrates efficient data handling strategies using multi-process parallel processing to minimize latency. Experimental results demonstrate that our optimized solution achieves up to an 8.96x improvement in inference speed compared to standard methods, while maintaining competitive performance.
\end{abstract}

%%%%%%%%% BODY TEXT
\section{Introduction}
In recent years, the rapid development of large-scale pre-trained language models \cite{yang2019semi, yang2019semi1, yang2018complex, yang2019deep, yang2021corporate, yang2024robust,li2024map,yang2018semi} based on the transformer architecture has led to significant breakthroughs in natural language processing (NLP) generation tasks \cite{fu2024noise, wancovlr, yang2021corporate, yang2021learning, wancovlr, yang2021cost, yang2021s2osc,yang2016learning,yang2015auxiliary}. By the end of 2022, ChatGPT \cite{floridi2020gpt,achiam2023gpt} gained immense popularity online, amassing over 100 million monthly active users in a short period. Its conversational quality approached or even surpassed human levels, establishing it as one of the most sought-after technologies.

Against this backdrop, leveraging deep AI and AIGC technological expertise, Baidu's commercial engine deployed the Ernie Wenxin model \cite{zhang2021ernie} to notably enhance marketing effectiveness for clients. However, in the practical deployment of commercial Ernie models, optimizing high-performance inference for complex generation models is crucial. This optimization ensures both seamless customer marketing experiences and effective control of computational costs. Therefore, GPU acceleration technology \cite{owens2008gpu} has emerged as a pivotal approach to enhancing inference performance. Across the industry, numerous outstanding technical solutions continue to emerge. Our objective is to employ typical Ernie generation models and various optimization techniques to challenge and achieve optimal inference performance.

We leverage the Baidu Paddle inference framework \cite{thiagarajan2018paddle} to creatively introduce more efficient acceleration methods for inference, building upon established technologies. We apply a series of mature technologies, including Faster Transformer \cite{vyas2020fast} and Faster Tokenizer \cite{song2020fast} for high-performance inference, FP16 half-precision inference \cite{haidar2018harnessing}, and efficient inference framework technologies based on the Paddle inference framework \cite{thiagarajan2018paddle}. Additionally, more innovative solutions are proposed to further enhance inference performance. At the model level, we effectively reduced computational costs by trimming the embedding layer. For data processing, we optimized the allocation of data inference order and extracted relevant content offline to minimize inefficient inference overhead. On the optimization front, we employed multi-process parallel processing to fully and efficiently utilize the computational performance of CPUs and GPUs. In particular, our experiments showcase that our solution not only accelerates inference but also maintains high levels of performance comparable to standard precision models.

The main contributions can be summarized as follows:

$\bullet$ For model-level optimization, we adopted the high-performance inference engine Faster Transformer, embedding layer pruning techniques, and FP16 half-precision inference, effectively reducing inference overhead.

$\bullet$ For processing optimization, we specifically enhanced the data handling capabilities of large language models. We leveraged the high-performance inference framework Paddle Inference and employed multi-process parallel processing to minimize data processing latency as much as possible.

$\bullet$ Our solution fully exploits the potential of the hardware and significantly reduces inference overhead. Extensive experiments validated the effectiveness of our proposal, optimizing baseline inference performance by approximately 8.96 times.

\section{Related Work}
\subsection{Ernie Large-scale Language Model}
Ernie model's \cite{zhang2021ernie} internal architecture is primarily based on the Transformer framework, a revolutionary deep learning model well-suited for processing sequential data such as natural language text \cite{nadkarni2011natural}. Here are the key principles of Ernie model's internal workings:

\textbf{Transformer Architecture}: Ernie is based on the Transformer model, which employs self-attention mechanisms to capture dependencies between different positions in the input sequence, mitigating the long-range dependency issues present in traditional recurrent neural networks. The Transformer consists of encoder and decoder layers, each equipped with multi-head attention mechanisms and feed-forward neural networks.

\textbf{Knowledge Integration}: Ernie enhances text representation by integrating rich linguistic knowledge, encompassing language contexts, syntactic rules, semantic relationships, and other hierarchical language structures. This integration enables the model to accurately comprehend and generate natural language.

\textbf{Multi-task Learning}: Ernie supports multi-task learning, enabling a single model to simultaneously handle multiple natural language processing tasks such as text classification, named entity recognition, sentiment analysis, etc. This approach enhances model generalization and efficiency.

\subsection{Model Optimization}

\textbf{Faster Transformer.} The Faster Transformer architecture introduces optimizations to the traditional transformer model, such as kernel fusion \cite{wang2010kernel} and tensorization techniques \cite{garipov2016ultimate}. These optimizations reduce memory bandwidth requirements and enhance parallelism, resulting in accelerated inference speeds for sequence-to-sequence tasks in natural language processing and machine translation \cite{koehn2009statistical}.

\textbf{Model Compression.} Model compression techniques \cite{buciluǎ2006model}, such as quantization \cite{gray1998quantization} and pruning \cite{liu2018rethinking}, aim to reduce the computational complexity of neural networks without compromising accuracy. These methods are crucial for optimizing inference performance by minimizing memory footprint and computational overhead.

\textbf{Key-Value (KV) Cache.} KV cache mechanisms store intermediate results of key computations, reducing redundant calculations during inference. This approach optimizes memory usage and speeds up inference by facilitating quicker access to frequently accessed data.

\subsection{Processing Optimization}

\textbf{Dynamic Batch Size.} Dynamic batch size strategies \cite{das2014adaptive} dynamically adjust batch sizes during inference based on available computational resources and input data characteristics. This approach optimizes GPU utilization by efficiently handling varying input sizes, thereby improving inference efficiency without sacrificing throughput or accuracy.

\textbf{Faster Tokenizer Faster.} Tokenizer techniques \cite{\cite{song2020fast} } focus on optimizing tokenization processes by leveraging efficient data structures and algorithms. These methods accelerate preprocessing steps, reducing overall inference time for transformer-based models like BERT and GPT \cite{floridi2020gpt}.

\textbf{Parallel Inference.} Parallelizing inference tasks \cite{yan2009parallel} across multiple processors or GPUs enhances throughput and reduces latency. Techniques like model parallelism and data parallelism enable simultaneous computation of different parts of a neural network, leveraging hardware capabilities for faster inference.

\textbf{Multi-Process Handling.} Utilizing multiple processes \cite{hintz1991multi} for inference enables concurrent execution of multiple model instances or tasks, improving overall system throughput. This approach is effective in distributed environments where workload distribution and resource utilization are critical for optimizing performance.
%-------------------------------------------------------------------------

% \begin{figure*}
% \begin{center}
% \includegraphics[width=\linewidth]{unimo.png}
% \end{center}
%    \caption{Overall Architecture. On the left of the figure is the pedestrian model architecture. On the basis of IRRA, Attribute Classification, and Inclusion Relation Matching are innovatively added. The design of the model fully mines the data annotation information and makes the performance of image-text pairs more excellent in fine-grained matching. The right side of the figure shows the vehicle model architecture. We adopt BLIP as a framework to optimize the fine-grained matching of image-text pairs and mine difficult samples. We also introduce a visual prompt augmentation strategy to give the model prior knowledge to help the model better recognize the vehicle color. }
% \label{fig:short}
% \end{figure*}

\begin{figure}[t]
	\centering
	\includegraphics[width=0.7\linewidth]{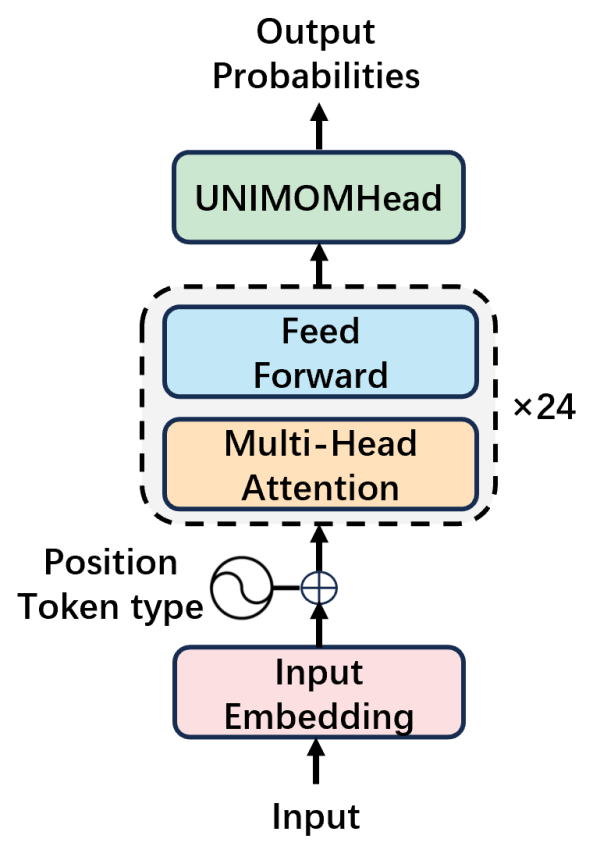}
	\caption{unimo architecture.}\label{fig:unimo}
\end{figure}

\begin{figure}[b]
	\centering
	\includegraphics[width=0.9\linewidth]{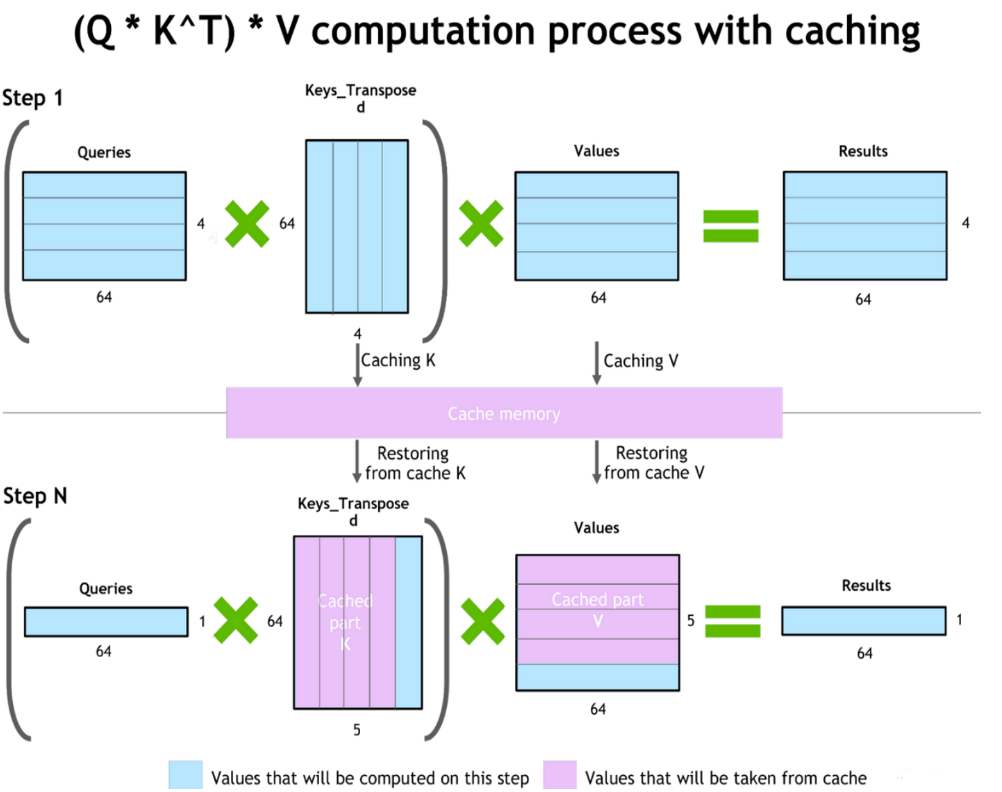}
	\caption{K-V cache.}\label{fig:faster_k_v}
\end{figure}

\begin{figure}[b]
	\centering
	\includegraphics[width=1.0\linewidth]{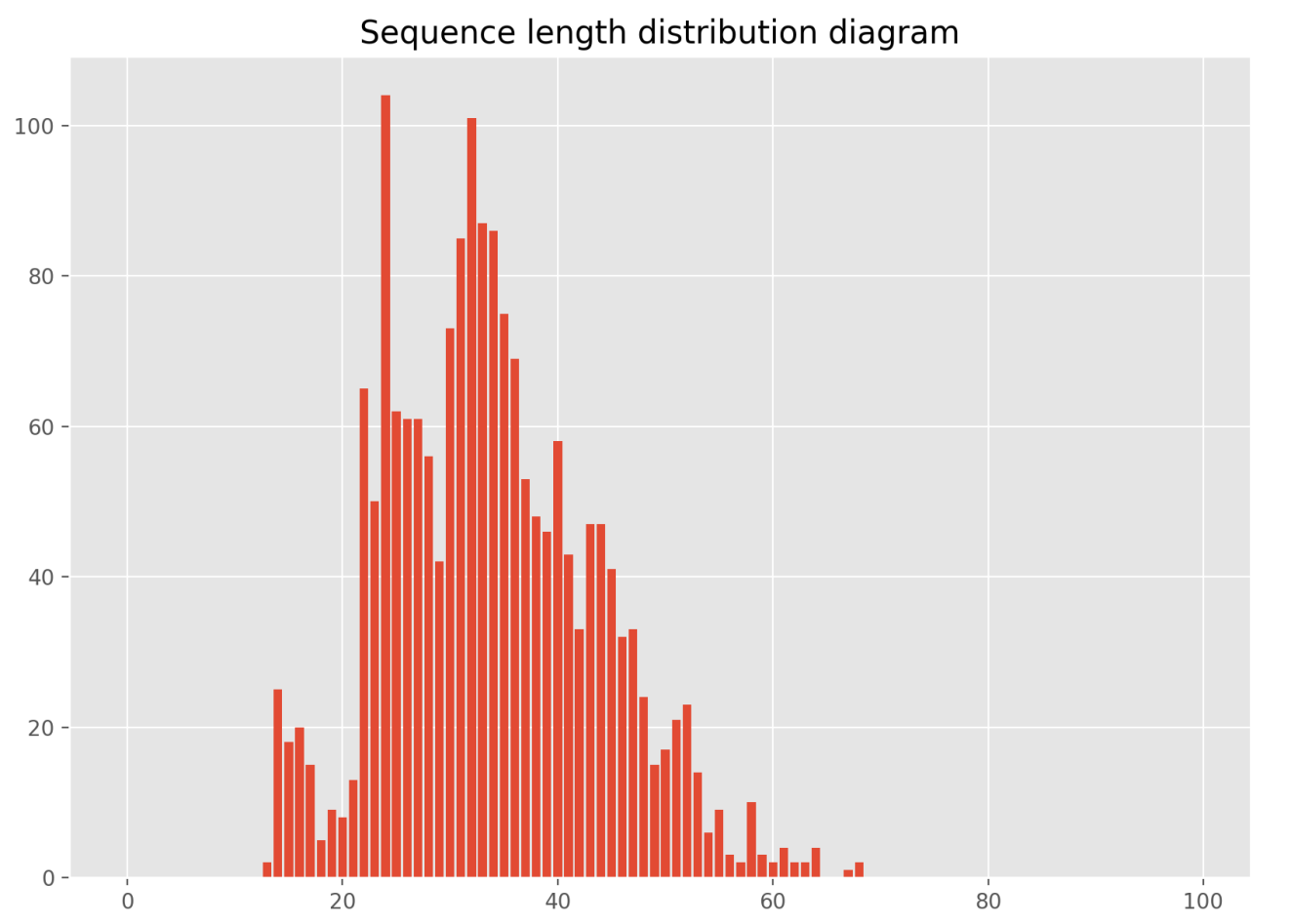}
	\caption{Sequence length.}\label{fig:seq_len}
\end{figure}

\begin{figure}[t]
	\centering
	\includegraphics[width=1.0\linewidth]{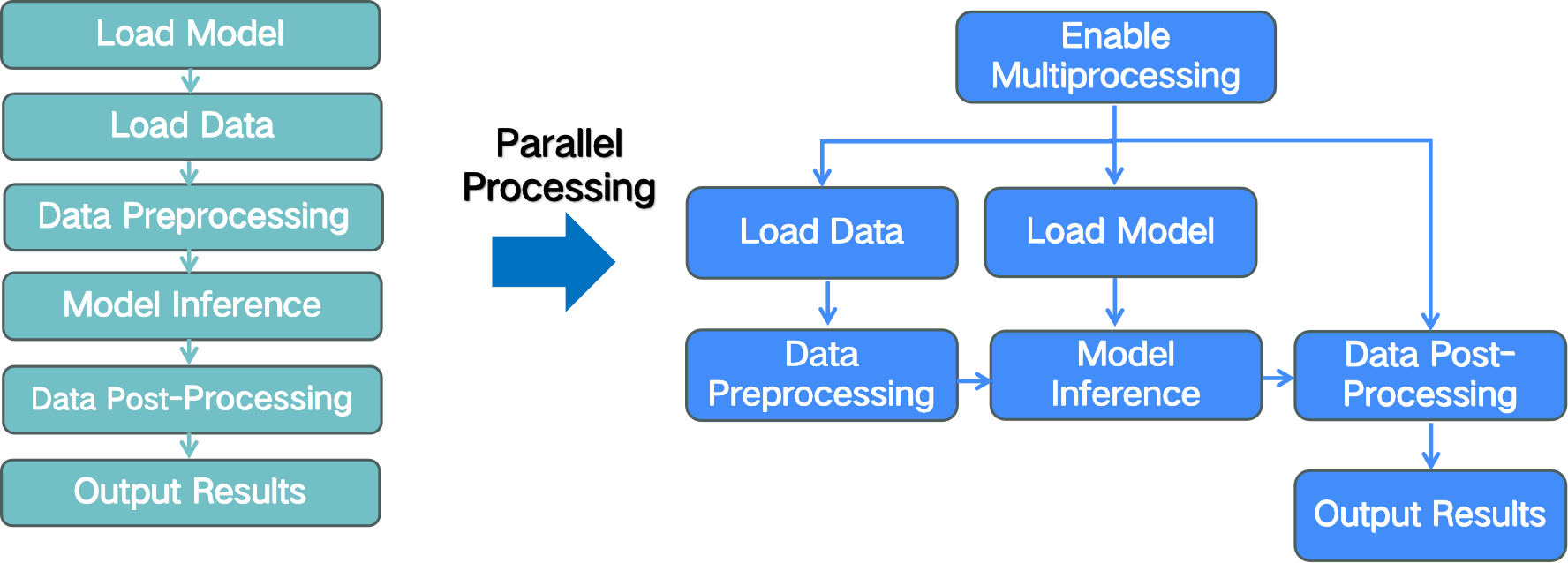}
	\caption{Multi-process parallel processing.}\label{fig:multiprocessing}
\end{figure}

\section{Method}
\subsection{Overall Architecture}

As shown in the Figure \ref{fig:unimo}, our core method is based on the pretrained multimodal model UNIMO-text. This model integrates text and image data to learn various types of information, significantly enhancing the model's generalization capability. The model structure includes 24 layers of stacked networks, which help to effectively handle complex input patterns and relationships. By training on large-scale datasets, this architecture optimizes its weights to achieve outstanding performance in various specific tasks.

%-------------------------------------------------------------------------
\subsection{Model-level optimization}
\textbf{Faster Transformer.} In our research, we have strategically chosen to integrate the Faster Transformer, a decision that significantly bolsters inference velocity without undermining the model's effectiveness. This enhancement is achieved through three pivotal technologies:

\begin{enumerate}
    \item \textbf{Precision Reduction to FP16:} By adopting a lower computational precision, we manage to preserve robust performance across various applications. This approach notably accelerates processing speeds and curtails memory demands, maintaining efficiency without compromising output quality.

    \item \textbf{Optimization of Matrix Multiplication:} Central to boosting computational throughput is the refinement of each phase in GEMM (General Matrix Multiply). This meticulous optimization of matrix operations expedites both forward and backward propagation, a benefit that becomes particularly salient when managing extensive datasets.
    
    \item \textbf{Key-Value (K-V) Cache Mechanism:} The implementation of a caching mechanism for key (K) and value (V) pairs substantially elevates the efficiency of recurrent computations, a feature prominently advantageous in the self-attention mechanism of the model. By storing previously computed key and value pairs, the model minimizes data reloading and computational redundancies during continuous inference phases. As shown in the Figure \ref{fig:faster_k_v}, this results in direct access to these matrices from the cache subsequent to their initial computation, thereby eliminating superfluous recalculations and markedly improving operational efficiency.
\end{enumerate}

These technological advancements collectively contribute to a transformation that not only accelerates inference but also preserves, if not enhances, the overall performance of the model.

%-------------------------------------------------------------------------
\textbf{Embedding layer pruning.} We conducted an in-depth analysis of the embedding layer structure of the UNIMO model. The embedding layer of the UNIMO model has 12800 dimensions, which plays a fundamental role in processing both Chinese and English. However, we observed that the embedding layer contains a large number of rarely used characters, which are seldom utilized during the model's inference process. To address this, we trimmed the vocabulary, retaining only high-frequency words to reduce the size of the word embedding matrix, thereby enhancing the model's inference speed. The fundamental architecture of the UNIMO model is built upon the Transformer.

In the Transformer architecture, adding positional information to the model inputs through the position embedding matrix is crucial. Positional embeddings enable the model to understand the relative or absolute positions of words in the text, which is vital for handling the sequence dependencies in natural language processing. In the UNIMO model, positional embeddings are expressed with a fixed dimension, with the embedding matrix size being 512×1024. As shown in the Figure \ref{fig:seq_len}, in real-world applications, the length of input sentences is typically less than 100 words, leading to a significant waste of computational resources. Therefore, we trimmed the position embedding matrix from 512×1024 to 128×1024 to optimize computational efficiency.

%------------------------------------------------------------------------
\subsection{Processing Optimization}

\textbf{Paddle Inference Optimization Techniques.} Paddle Inference is the dedicated inference engine of the PaddlePaddle deep learning framework developed by Baidu. It focuses on enhancing model inference efficiency and performance. The primary optimization techniques include:

\begin{enumerate}
    \item \textbf{Memory Reuse to Increase Throughput:}
    Memory reuse technology optimizes the utilization of memory resources to enhance overall inference throughput. This involves reusing the memory space of mutually independent tensors, allowing multiple computation tasks to run in parallel without the need for separate memory allocations for each task. This approach not only reduces overall memory usage but also significantly increases computational parallelism, thus enhancing the throughput of the inference process. In summary, memory reuse technology facilitates faster inference speeds and greater data processing capacity through more efficient resource utilization.

    \item \textbf{Fine-Grained OP Horizontal and Vertical Fusion to Reduce Computation:}
    Fine-grained operator (OP) fusion, both horizontal and vertical, reduces computation and improves efficiency by consolidating multiple operations into a single operation. The specific methods include:
    \begin{itemize}
        \item \textbf{Horizontal Fusion:} Merges adjacent operations of the same type, for instance, combining consecutive addition operations into one operation.
        \item \textbf{Vertical Fusion:} Combines different types of operations that can be executed jointly, such as merging an addition operation with a multiplication operation into a composite operation.
    \end{itemize}
    This approach not only reduces the total number of operations within the model, thus decreasing computation, but also lowers the number of kernel launches. A kernel launch, which involves initiating a computational kernel on the GPU, incurs some overhead. Reducing the number of kernel launches can further enhance inference performance.
\end{enumerate}

%------------------------------------------------------------------------
\textbf{Multi-process parallel processing.} The traditional model inference process includes the following steps: model loading, data loading, data preprocessing, model inference, data post-processing, and result output. These steps usually need to be executed sequentially in order. However, we have observed that these different process steps can be handled in parallel using multi-process techniques. As shown in the Figure \ref{fig:multiprocessing}, we divide the entire process into four main processes using multi-process parallel processing technology: the main process, the data preprocessing process, the model inference process, and the data post-processing process. This method can significantly improve processing efficiency.

%------------------------------------------------------------------------
\section{Experiments}
\textbf{Dataset.} The dataset used in this study is sourced from Baidu's commercial material data, intended to provide textual materials for training and validating NLP models. The dataset is loaded using the \texttt{load\_dataset} API provided by PaddleNLP, including approximately 2,000 test dataset samples, 10,000 regional competition validation dataset samples, and 50,000 semifinal competition validation dataset samples. This dataset is specifically designed for tasks involving text summarization and content understanding. The primary difference between the test and validation datasets is that the validation dataset lacks the \texttt{summary} field, requiring the model to generate corresponding summaries automatically.

\textbf{Result.} Table \ref{table:1} presents the performance metrics for various methods applied to enhance speed in our testing setup. The methods were introduced sequentially, as indicated by the order in the table. Our experimental outcomes robustly validate the efficacy of our proposed techniques. Initially, employing a baseline method achieved a speed of 16.11. The introduction of the Fast Transformer significantly boosted performance to 98.46, demonstrating its capability to expedite processing. Further enhancements were achieved with the embedding layer pruning, which raised the speed to 125.32, and the implementation of multi-process parallel processing, which culminated in the top speed of 144.45. The consistent increase in speed with each method addition underlines the effectiveness of our integrated approach in optimizing computational efficiency.

\begin{table}[!htbp]
\caption{Results of our method on the test set.}
\label{table:1}
\begin{center}
\begin{tabular}{ >{\centering\arraybackslash}m{0.05cm} 
>{\centering\arraybackslash}m{6.1cm} 
>{\centering\arraybackslash}m{0.7cm} 
}
 \hline
 \# & Method &  Speed \\ 
 \hline
 1 & Baseline & $ 16.11 $\\ 
 2 & Fast transformer & $ 98.46 $\\ 
 3 & embedding layer pruning & $ 125.32 $\\ 
 4 & multi-process parallel processing & $ 144.45 $ \\
 \textbf{5} & \textbf{Final result} & $\textbf{144.45}$ \\ 
\hline
\end{tabular}
\end{center}
\end{table}

\section{Conclusion}
This paper details optimization strategies for the UNIMO model to enhance its inference performance, using the Paddle inference framework. Techniques like the Faster Transformer, embedding layer pruning, and FP16 half-precision inference significantly increase computational efficiency. Additionally, multi-process parallel processing is utilized to minimize latency and reduce computational overhead. These improvements allow our approach to achieve an 8.96-fold increase in inference speed while maintaining high performance, demonstrating the potential for practical AI applications in commercial environments. Notably, our methods secured the first-place position in the semifinal evaluations.

{\small
\bibliographystyle{ieee}
\bibliography{egpaper_final}
}

\end{document}